\pdfoutput=1

\documentclass[11pt]{article}

\usepackage[final]{acl}

\usepackage{times}
\usepackage{latexsym}

\usepackage[T1]{fontenc}

\usepackage[utf8]{inputenc}

\usepackage{microtype}

\usepackage{inconsolata}

\usepackage{graphicx}

\usepackage{graphicx}
\usepackage{subcaption}
\usepackage{multirow}
\usepackage{booktabs}

\title{A Systematic Study of Cross-Layer KV Sharing \\ for Efficient LLM Inference}

\author{You Wu\thanks{\; Equal contribution.}, Haoyi Wu\footnotemark[1], Kewei Tu\thanks{\; Corresponding author.} \\
School of Information Science and Technology, ShanghaiTech University \\
Shanghai Engineering Research Center of Intelligent Vision and Imaging \\
\texttt{\{wuyou2024, wuhy1, tukw\}@shanghaitech.edu.cn}}

\begin{document}
\maketitle
\begin{abstract}

Recently, sharing key-value (KV) cache across layers has been found effective in efficient inference of large language models (LLMs).
To systematically investigate different techniques of cross-layer KV sharing, we propose a unified framework that covers several recent methods and their novel variants.
We conduct comprehensive experiments on all the configurations of the framework, evaluating their generation throughput and performance in language modeling and downstream tasks.
We find that when reducing the size of the KV cache by $2\times$, most configurations can achieve higher throughput than standard transformers while maintaining competitive performance.
When further reducing the size of the KV cache, however, pairing queries of all layers with KVs of upper layers performs better, at the expense of additional training cost and prefilling latency.
We hope that this work will help users make more informed choices of cross-layer KV sharing approaches and facilitate future research on efficient LLM inference.

\end{abstract}

\section{Introduction}

A major bottleneck for the deployment of LLMs is memory consumption, of which the key-value (KV) cache in the transformer architecture occupies a large portion \cite{kwon2023efficient}.
Various methods have been proposed to reduce the memory consumption of the KV cache in LLMs.
For example, \citet{shazeer2019fast, ainslie-etal-2023-gqa} share the KVs across query heads and \citet{zhang2023ho, xiao2024efficient} keep the KV cache of only a small portion of tokens. 

More recently, several methods are proposed in which the KVs are computed only at a subset of transformer layers and shared to the other layers, such as LCKV \cite{wu-tu-2024-layer}, YOCO \cite{sun2024you} and CLA \cite{brandon2024reducing}.
These methods not only significantly reduce memory consumption but also improve inference speed, while preserving the performance of LLMs in language modeling and downstream tasks.
However, while all these methods are based on the idea of cross-layer KV sharing, they differ significantly in how the sharing is done.

In this study, we consider a unified framework for cross-layer KV sharing, of which LCKV, CLA, and YOCO can be seen as special configurations. 
We then empirically test all the configurations of the framework, including several novel ones that have never been considered in previous work.
Our experiments show that, with respect to throughput, all the configurations can achieve significantly higher throughput than the standard transformer when the prompt is short; but when the prompt is long, the throughput of the configurations that compute the KVs at the top layers degrades dramatically.
With respect to performance, when only half of the layers rely on the KVs computed by the other layers, the performance of most configurations is comparable with that of the standard transformer; when more layers become reliant on the other layers for the KVs, the configurations that compute the KVs at the bottom layers suffer the greatest performance degradation.
We hope our framework and empirical studies would help users interested in cross-layer KV sharing to make more informed choices of methods and configurations according to their throughput and performance requirements.
Our code is available at \url{https://github.com/whyNLP/LCKV}.

\section{Existing Methods}

Layer-Condensed KV Cache (LCKV) \cite{wu-tu-2024-layer} computes the KVs of only the top layer of the transformer, which are paired with queries of all the layers. Consequently, LCKV omits the KV computation and discards the KV parameters for all the layers other than the top layer.
To prevent severe performance degradation, LCKV also optionally retains standard attention for a small number of top and bottom layers.

You Only Cache Once (YOCO) \cite{sun2024you} computes the KVs of only the middle layer of the transformer, which are paired with the queries of the top-half of the layers.
The bottom-half of the layers uses efficient attention to achieve a constant cache size.
\citet{goldstein2024goldfinch} uses a similar sharing pattern to YOCO, but further compresses the size of the KV cache.

Cross-Layer Attention (CLA) \cite{brandon2024reducing} uniformly divides transformer layers into multiple groups of adjacent layers. In each group, it pairs the queries of all the layers with the KVs of the bottom layer.
\citet{zuhri2024mlkv} shares the KVs in the same way as CLA, but applies a more efficient training scheme.
\citet{liu2024minicache} groups every two adjacent layers in the middle-to-deep portion and compresses the KV cache in each group.
\citet{chen2024skip} groups non-adjacent layers and pairs the queries of the upper layer with the KVs of the lower layer in each group.
\citet{rajput2024inference} uses a combination of the sliding window attention and a sharing pattern similar to CLA.
\citet{liao2024beyond, mu2024cross, rajabzadeh2024echoatt} apply sharing patterns similar to CLA to the computed attention weights instead of KVs.

\section{A Unified Framework}

Unifying previous methods, we propose a framework for cross-layer KV sharing that can be applied to any transformer-based model. Suppose that the transformer has $L$ layers.
We denote $kv(i) \in \{1, ..., L\}$ as the index of the layer whose KVs are paired with the queries of the $i$-th layer.
If $ kv(i) = i $, then layer $i$ is called a \emph{KV layer}, which computes its own KVs that are paired with its queries just as in a standard transformer.
Otherwise, layer $i$ does not compute its own KVs and instead uses the KV of layer $ kv(i) \neq i$.
In this case, we call layer $kv(i)$ the \emph{target layer} of layer $i$.
Since layer $i$ does not need to compute KVs, it does not need weights $W_K, W_V$.
Therefore, the number of KV layers determines the number of weight parameters $W_K, W_V$ and hence the size of a transformer model.
Below we define different configurations of our framework assuming the number of KV layers always set to $l$.

We define a configuration by partitioning transformer layers and positioning target layer(s) differently.
We choose the layer partitioning from \{\emph{ pizza, sandwich, lasagna }\} and choose the target layer positioning from \{\emph{ bottom, top, middle }\}\footnote{We also consider positioning at quarter and three-quarter, which is discussed in Appendix \ref{apx:quarter}.}.
The pizza partitioning sets the first $l-1$ layers as KV layers.
The sandwich partitioning sets the first $\lceil \frac{l-1}{2} \rceil$ layers and the last $\lfloor \frac{l-1}{2} \rfloor$ layers as KV layers.
For the remaining $L-l+1$ consecutive layers in both pizza and sandwich, their target layer is positioned at either the top, the middle, or the bottom of these layers.
The lasagna partitioning uniformly divides the $L$ layers into $l$ groups of consecutive layers.
For each group except the first, the target layer of all the layers within the group is positioned at either the top, the middle, or the bottom of these layers.
For the first group, however, we always set the bottom layer as the target layer because we empirically find that there is a significant drop in performance if the first layer is not a KV layer.

Note that for the top and middle positioning of the target layer, there exists a cyclic dependency between the target layer and the lower non-KV layers: for each token, its KVs at the target layer is required for attention computation at lower non-KV layers, but are not computed until computation at all the lower layers is finished. So, we follow \citet{wu-tu-2024-layer} and drop the attention of each token to itself, which is equivalent to masking the diagonal of the attention matrix in each layer.

Table \ref{tab:model} illustrates all the nine configurations that we have defined.
We name each configuration with its partitioning and positioning pattern.
The sandwich-top, pizza-bottom and lasagna-bottom configurations correspond to LCKV, YOCO\footnote{The pizza-bottom configuration differs from YOCO in that it uses the standard attention instead of the efficient attention for the bottom-half of the layers.} and CLA respectively.
The lasagna-top configuration and all middle configurations are novel and have not been considered in previous work.

\begin{table}[tbh]
  \centering
  \small
  \setlength\tabcolsep{2pt}
  \begin{tabular}{|c|c|c|c|c|}
    \hline
    \multicolumn{2}{|c|}{} & \multicolumn{3}{c|}{Layer Partitioning} \\
    \cline{3-5}
    \multicolumn{2}{|c|}{} & Pizza & Sandwich & Lasagna \\
    \hline
    \multirow{3}{*}{\rotatebox{90}{Target Layer Positioning \hspace{2.75cm}}} & \rotatebox[origin=c]{90}{Bottom} &
    \begin{minipage}[b]{0.25\columnwidth}
        \centering
        \vspace{0.2ex}
        \raisebox{-.5\height}{\includegraphics[trim = 125mm 55mm 125mm 55mm, clip, width=\columnwidth]{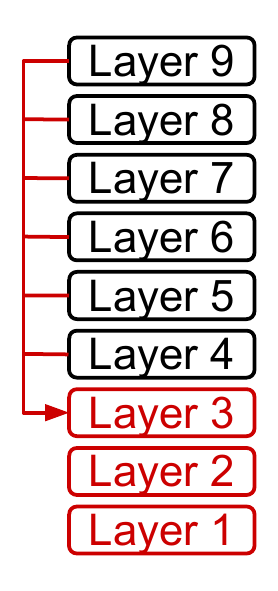}}
    \end{minipage} &
    \begin{minipage}[b]{0.25\columnwidth}
        \centering
        \vspace{0.2ex}
        \raisebox{-.5\height}{\includegraphics[trim = 125mm 55mm 125mm 55mm, clip, width=\columnwidth]{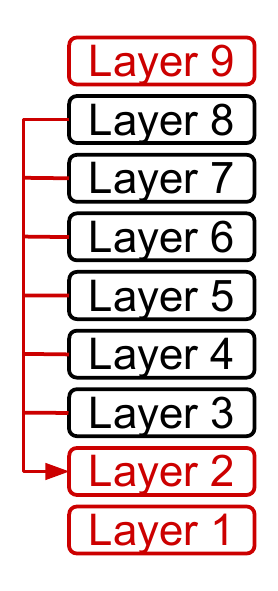}}
    \end{minipage} &
    \begin{minipage}[b]{0.25\columnwidth}
        \centering
        \vspace{0.2ex}
        \raisebox{-.5\height}{\includegraphics[trim = 125mm 55mm 125mm 55mm, clip, width=\columnwidth]{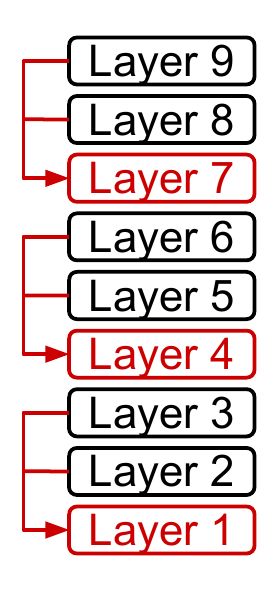}}
    \end{minipage} \\
    \cline{2-5}
    & \rotatebox[origin=c]{90}{Top} &
    \begin{minipage}[b]{0.25\columnwidth}
        \centering
        \vspace{0.2ex}
        \raisebox{-.5\height}{\includegraphics[trim = 125mm 55mm 125mm 55mm, clip, width=\columnwidth]{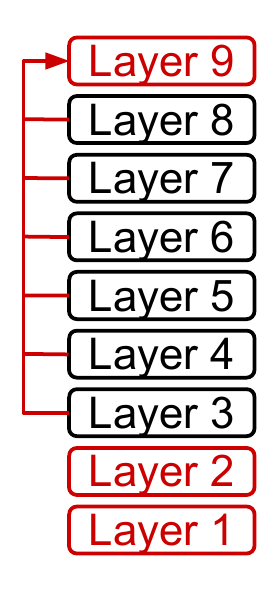}}
    \end{minipage} &
    \begin{minipage}[b]{0.25\columnwidth}
        \centering
        \vspace{0.2ex}
        \raisebox{-.5\height}{\includegraphics[trim = 125mm 55mm 125mm 55mm, clip, width=\columnwidth]{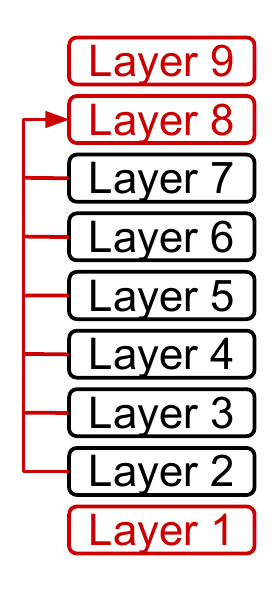}}
    \end{minipage} &
    \begin{minipage}[b]{0.25\columnwidth}
        \centering
        \vspace{0.2ex}
        \raisebox{-.5\height}{\includegraphics[trim = 125mm 55mm 125mm 55mm, clip, width=\columnwidth]{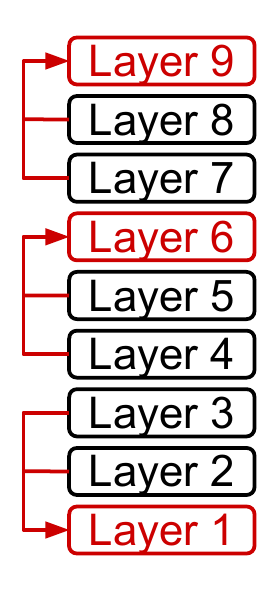}}
    \end{minipage} \\
    \cline{2-5}
    & \rotatebox[origin=c]{90}{Middle} &
    \begin{minipage}[b]{0.25\columnwidth}
        \centering
        \vspace{0.2ex}
        \raisebox{-.5\height}{\includegraphics[trim = 125mm 55mm 125mm 55mm, clip, width=\columnwidth]{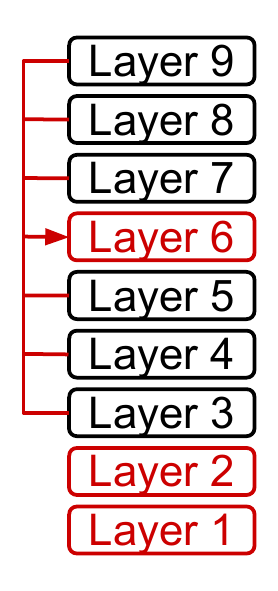}}
    \end{minipage} &
    \begin{minipage}[b]{0.25\columnwidth}
        \centering
        \vspace{0.2ex}
        \raisebox{-.5\height}{\includegraphics[trim = 125mm 55mm 125mm 55mm, clip, width=\columnwidth]{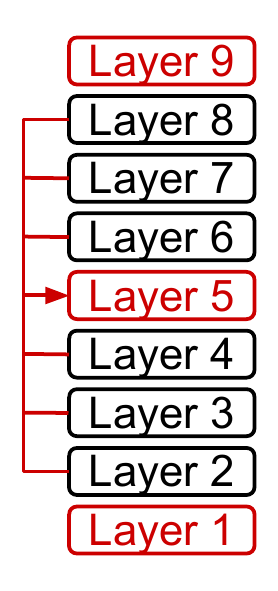}}
    \end{minipage} &
    \begin{minipage}[b]{0.25\columnwidth}
        \centering
        \vspace{0.2ex}
        \raisebox{-.5\height}{\includegraphics[trim = 125mm 55mm 125mm 55mm, clip, width=\columnwidth]{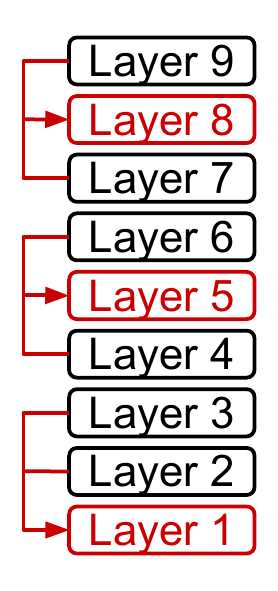}}
    \end{minipage} \\
    \hline
  \end{tabular}
  \caption{All the configurations in our unified framework for cross-layer KV sharing. Red layers are KV layers. Each arrow points to a target layer from the layers whose queries are paired with its KV. The sandwich-top configuration corresponds to LCKV, the pizza-bottom configuration corresponds to YOCO, and the lasagna-bottom configuration corresponds to CLA.}
  \label{tab:model}
\end{table}

\subsection{Training}

For the bottom positioning, the model can be trained in the same way as a standard transformer model.
For the top and middle positioning, however, the attention computation of each token at layer $i < kv(i)$ depends on KVs of the previous tokens at its target layer $kv(i)$, creating sequential dependencies that spoil parallel training. 
Following \citet{wu-tu-2024-layer}, we perform iterative training to break the sequential dependencies.
In each iteration, we pair the queries of each layer with the KVs of its target layer from the previous iteration.
For a token sequence of length $n$, parallel training with $n$ iterations is equivalent to sequential training.
In order to reduce the training cost, we backpropagate the loss only through the last $b$ iterations, and use $m \ll n - b$ iterations to approximate the KVs of the first $n - b$ iterations.

Note that not all layers need to be trained iteratively.
For some configurations, there exist layers without any sequential dependencies at the top and bottom, and we can compute these layers in one pass before and after iterative training, respectively.
Therefore, for the pizza and sandwich partitioning, we perform iterative training only on the layers ranging from the first non-KV layer to its target layer, and for the lasagna partitioning, we perform iterative training only on the layers ranging from the first layer of the second group and the target layer of the last group.

\subsection{Inference}

The inference of LLMs can be divided into the prefilling and decoding stages.
During the prefilling stage, we can conduct early exit \cite{sun2024you} after computing the KVs of the last KV layer.
For the top and middle positioning, we perform parallel encoding of the prompt in spite of sequential dependencies by iterative computation with $m + b$ iterations in the same way as in training.
The decoding stage is the same as in a standard transformer.

\section{Experiments}
\label{sec:experiments}

We conduct experiments to compare the generation throughput and performance of the standard Llama baseline \cite{touvron2023llama} and the nine configurations with different numbers of KV layers.
Our implementation is based on HuggingFace Transformers \cite{wolf-etal-2020-transformers} with kernel replacement with FlashAttention 2 \cite{dao2024flashattention}, fused RMS norm, fused cross-entropy, and fused SwiGLU.
Our experiments are conducted on models with 110M and 1.1B parameters, whose configurations are shown in Appendix \ref{apx:details}.
We set $m=7$ and $b=2$ for the top and middle configurations.
The sandwich configurations coincide with the pizza configurations when there are only two KV layers and the lasagna-middle configuration coincides with the lasagna-top configuration when the number of KV layers is half of the total number of layers (i.e., 6 and 11 for the 110M and 1.1B models, respectively), therefore omitted in our experiments.

\subsection{Generation Throughput}
\label{sec:throughput}

We test the generation throughput of the standard Llama and the nine configurations with 1.1B parameters on an RTX 3090 (24GB) GPU with different sequence lengths.
The evaluation follows the settings of FlexGen \cite{sheng2023flexgen}.

Figure \ref{fig:experiment}(a) reports the maximum throughput\footnote{The throughput at different batch sizes is shown in Appendix \ref{apx:batchsize}.}.
When the prompt is short (i.e., 5+2043), the prefilling time can be ignored and the generation throughputs of all the nine configurations are almost identical, which are much higher than the baseline throughput and increase as the number of KV layers decreases.
When the prompt is long (i.e., 512+1024), the prefilling time becomes significant for the top and middle configurations because of iterative encoding of the prompt. Consequently, their throughputs degrade dramatically, falling below the baseline in some cases.
On the other hand, the bottom configurations still achieve significantly higher throughputs than the baseline because no additional computation for prompt is required.

\subsection{Performance on Small Training Set}
\label{sec:small-experiments}

We train the standard Llama and the nine configurations with 110M and 1.1B parameters from scratch\footnote{We also tried model initialization with pre-trained models, the results of which are shown in Appendix \ref{apx:initialize}.} on the Minipile dataset \cite{kaddour2023minipile} with 1.7B tokens for one epoch and two epochs, respectively, and evaluate their perplexity. The training details are shown in Appendix \ref{apx:details}.

Figure \ref{fig:experiment}(b) reports the perplexity.
It can be seen that more KV layers lead to better performance in most cases.
When the number of KV layers is half of the total number of layers, the performance of most configurations is comparable with that of the baseline.
As we reduce the number of KV layers, the performance degrades for almost all the configurations, but the top and middle configurations are less affected compared to the bottom configurations. 
Two exceptions are the lasagna-top and lasagna-middle configurations, whose performance usually improves with fewer KV layers.
This may be due to the fact that the more KV layers there are, the more difficult it is to accurately approximate all the KVs with iterative training.

It can also be seen that the pizza-bottom and lasagna-bottom configurations perform relatively well among all the bottom configurations, and the sandwich-top and sandwich-middle configurations perform relatively well among all the top and middle configurations, respectively.
Therefore, we decide to train these four configurations with more data to further investigate their potential in language modeling and downstream tasks.

\subsection{Performance on Large Training Set}
\label{sec:large-experiments}

We train the standard Llama and the four well-performing configurations with 1.1B parameters from scratch on a 100B subset of the SlimPajama dataset \cite{cerebras2023slimpajama} for one epoch and evaluate their perplexity and downstream task accuracy. The training details are shown in Appendix \ref{apx:details}.
We evaluate the perplexity on a 10M subset of the development set of SlimPajama.
We also use the LM Eval Harness framework \cite{eval-harness} to test the zero-shot performance on commonsense reasoning tasks including Hellaswag \cite{zellers-etal-2019-hellaswag}, OpenBookQA \cite{mihaylov-etal-2018-suit}, WinoGrande \cite{sakaguchi2021winogrande}, ARC-Easy and ARC-Challenge \cite{clark2018think}, BoolQ \cite{clark-etal-2019-boolq}, PIQA \cite{bisk2020piqa}, and SciQ \cite{welbl-etal-2017-crowdsourcing}.

\begin{figure}[tbh]
    \centering
    \begin{subfigure}[b]{\columnwidth}
        \centering
        \includegraphics[trim = 0mm 140mm 0mm 0mm, clip, width=\columnwidth]{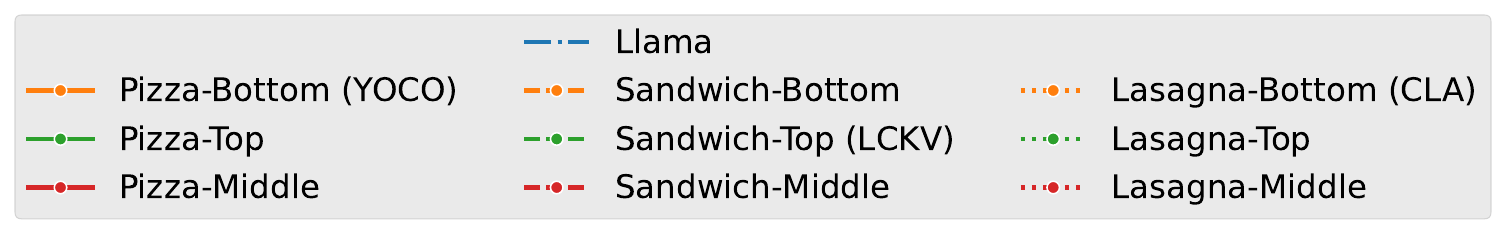}
    \end{subfigure}
    \begin{subfigure}[b]{\columnwidth}
        \centering
        \includegraphics[width=\columnwidth]{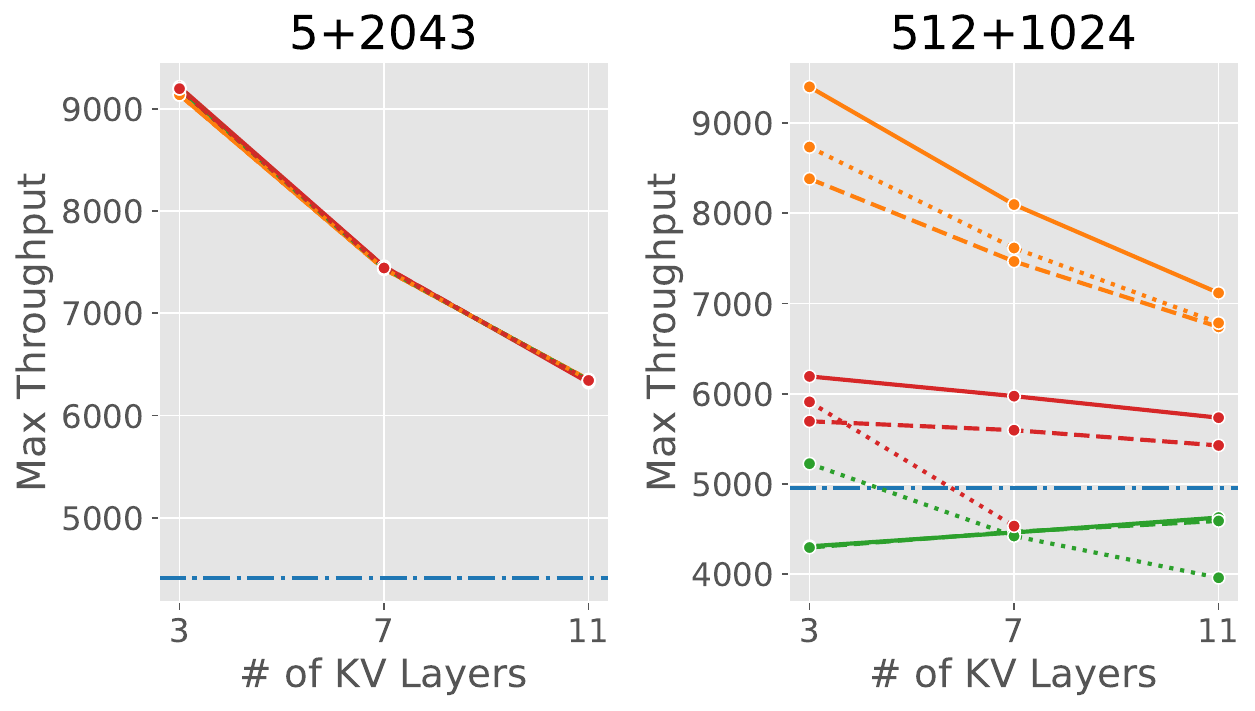}
        \caption{Maximum generation throughput on an RTX 3090 (24GB) GPU with different sequence lengths. We use ``$x + y$'' to denote a prompt length of $x$ and a generation length of $y$.}
    \end{subfigure}
    \begin{subfigure}[b]{\columnwidth}
        \centering
        \vspace{2mm}
        \includegraphics[width=\columnwidth]{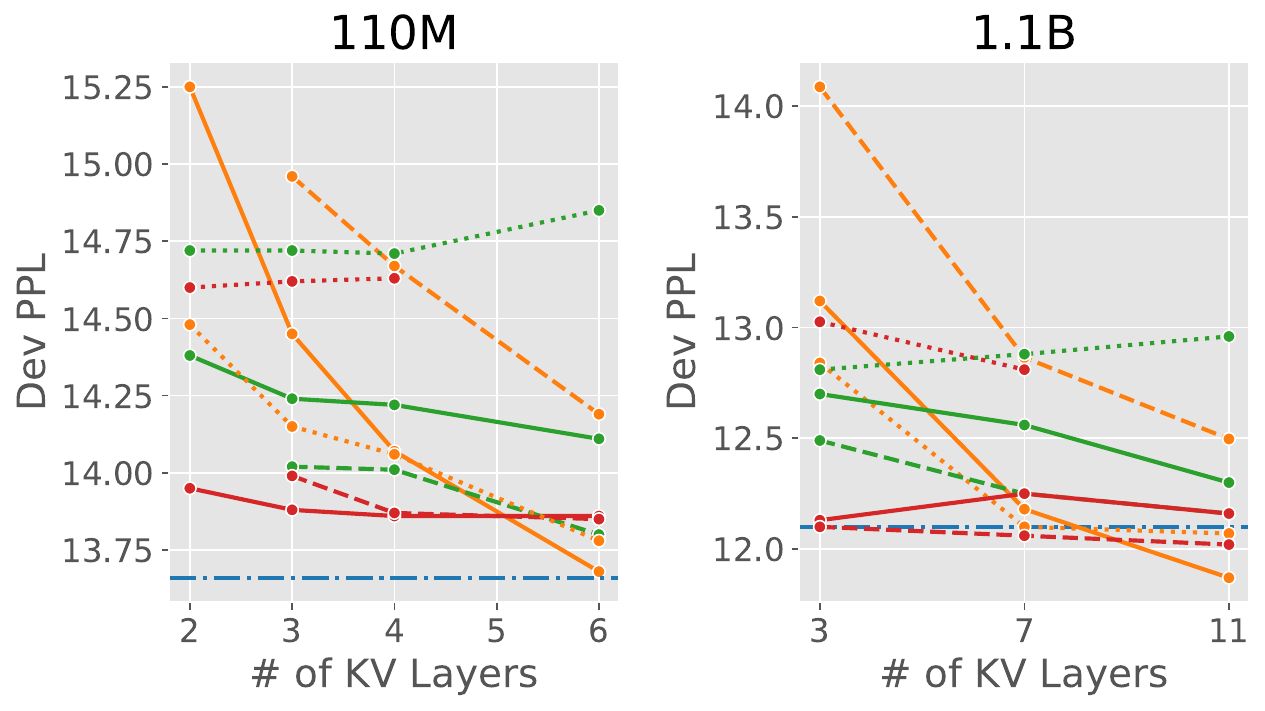}
        \caption{Perplexity on the Minipile dataset.}
    \end{subfigure}
    \begin{subfigure}[b]{\columnwidth}
        \centering
        \vspace{2mm}
        \includegraphics[width=\columnwidth]{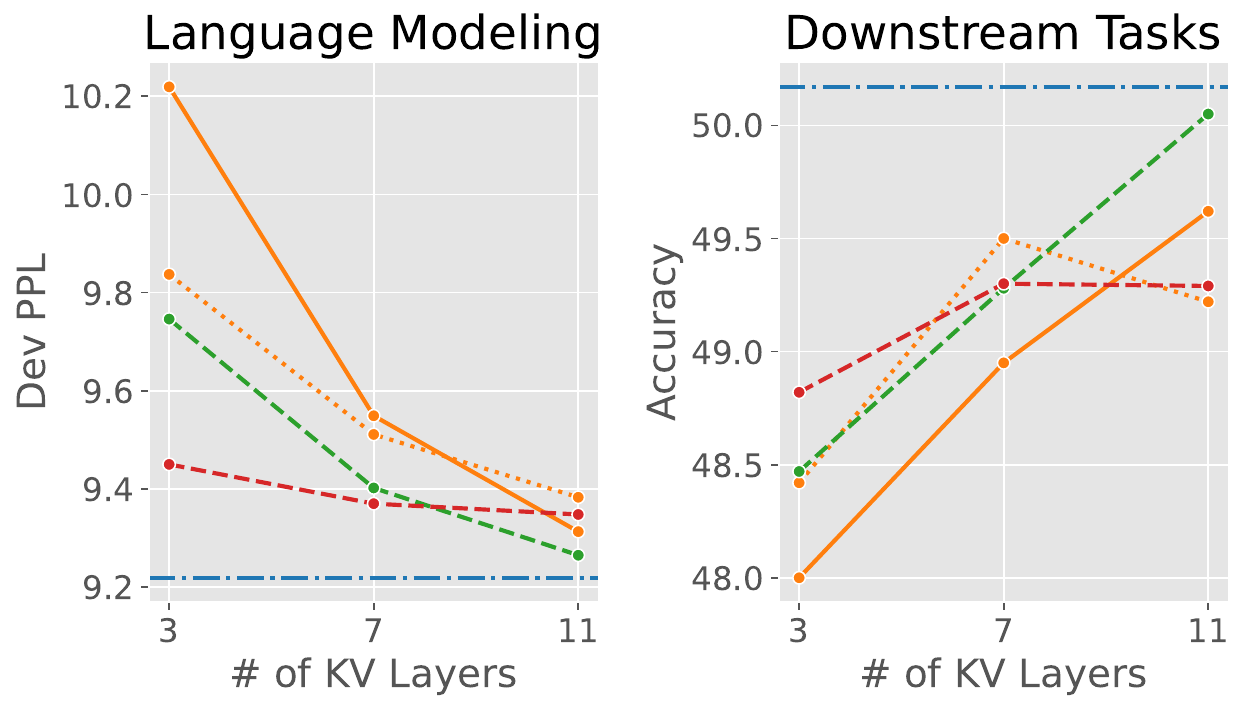}
        \caption{Perplexity on the SlimPajama dataset and downstream task results of 1.1B models.}
    \end{subfigure}
    \caption{Experimental results.}
    \label{fig:experiment}
\end{figure}

Figure \ref{fig:experiment}(c) reports the perplexity and average accuracy of downstream tasks.
Detailed results of downstream tasks are shown in Appendix \ref{apx:downstream}.
It can be seen that the sandwich-top configuration performs better than the two bottom configurations in both perplexity and downstream task accuracy, except for an outlier of the lasagna-bottom configuration with 7 KV layers in downstream task accuracy.
The sandwich-middle configuration performs best when the number of KV layers is small.

\section{Conclusion}

In this study, we propose a new framework for LLM cross-layer KV sharing that includes previous methods as special cases.
We conduct systematic experiments on various configurations of the framework with different KV cache memory budgets and observe their generation throughput and performance in language modeling and downstream tasks.
The experimental results show that the pizza-bottom and lasagna-bottom configurations can reduce the size of the KV cache by $2\times$ without too much performance degradation or introducing additional training and prefilling time.
However, if one wishes to further reduce the size of the KV cache, cares less about additional training time, and needs to generate sequences much longer than prompts, then the sandwich-middle configuration may be a better choice.

\section*{Limitations}

In this study, we only conduct experiments on models with 1.1B parameters and training set with 100B tokens. Due to the limited computational resources, we do not explore the performance of larger models with more training data.

\section*{Acknowledgements}

This work was supported by HPC Platform of ShanghaiTech University.

\bibliography{custom}

\begin{thebibliography}{31}
\providecommand{\natexlab}[1]{#1}

\bibitem[{Ainslie et~al.(2023)Ainslie, Lee-Thorp, de~Jong, Zemlyanskiy, Lebron, and Sanghai}]{ainslie-etal-2023-gqa}
Joshua Ainslie, James Lee-Thorp, Michiel de~Jong, Yury Zemlyanskiy, Federico Lebron, and Sumit Sanghai. 2023.
\newblock \href {https://doi.org/10.18653/v1/2023.emnlp-main.298} {{GQA}: Training generalized multi-query transformer models from multi-head checkpoints}.
\newblock In \emph{Proceedings of the 2023 Conference on Empirical Methods in Natural Language Processing}, pages 4895--4901, Singapore. Association for Computational Linguistics.

\bibitem[{Bisk et~al.(2020)Bisk, Zellers, Gao, Choi et~al.}]{bisk2020piqa}
Yonatan Bisk, Rowan Zellers, Jianfeng Gao, Yejin Choi, et~al. 2020.
\newblock Piqa: Reasoning about physical commonsense in natural language.
\newblock In \emph{Proceedings of the AAAI conference on artificial intelligence}, volume~34, pages 7432--7439.

\bibitem[{Brandon et~al.(2024)Brandon, Mishra, Nrusimha, Panda, and Kelly}]{brandon2024reducing}
William Brandon, Mayank Mishra, Aniruddha Nrusimha, Rameswar Panda, and Jonathan~Ragan Kelly. 2024.
\newblock Reducing transformer key-value cache size with cross-layer attention.
\newblock \emph{arXiv preprint arXiv:2405.12981}.

\bibitem[{Chen et~al.(2024)Chen, Wang, Zhang, Zheng, Zhang, Deng, Yu, Liu, Ma, and Zhang}]{chen2024skip}
Qian Chen, Wen Wang, Qinglin Zhang, Siqi Zheng, Shiliang Zhang, Chong Deng, Hai Yu, Jiaqing Liu, Yukun Ma, and Chong Zhang. 2024.
\newblock Skip-layer attention: Bridging abstract and detailed dependencies in transformers.
\newblock \emph{arXiv preprint arXiv:2406.11274}.

\bibitem[{Clark et~al.(2019)Clark, Lee, Chang, Kwiatkowski, Collins, and Toutanova}]{clark-etal-2019-boolq}
Christopher Clark, Kenton Lee, Ming-Wei Chang, Tom Kwiatkowski, Michael Collins, and Kristina Toutanova. 2019.
\newblock \href {https://doi.org/10.18653/v1/N19-1300} {{B}ool{Q}: Exploring the surprising difficulty of natural yes/no questions}.
\newblock In \emph{Proceedings of the 2019 Conference of the North {A}merican Chapter of the Association for Computational Linguistics: Human Language Technologies, Volume 1 (Long and Short Papers)}, pages 2924--2936, Minneapolis, Minnesota. Association for Computational Linguistics.

\bibitem[{Clark et~al.(2018)Clark, Cowhey, Etzioni, Khot, Sabharwal, Schoenick, and Tafjord}]{clark2018think}
Peter Clark, Isaac Cowhey, Oren Etzioni, Tushar Khot, Ashish Sabharwal, Carissa Schoenick, and Oyvind Tafjord. 2018.
\newblock Think you have solved question answering? try arc, the ai2 reasoning challenge.
\newblock \emph{arXiv preprint arXiv:1803.05457}.

\bibitem[{Dao(2024)}]{dao2024flashattention}
Tri Dao. 2024.
\newblock \href {https://openreview.net/forum?id=mZn2Xyh9Ec} {Flashattention-2: Faster attention with better parallelism and work partitioning}.
\newblock In \emph{The Twelfth International Conference on Learning Representations}.

\bibitem[{Gao et~al.(2023)Gao, Tow, Abbasi, Biderman, Black, DiPofi, Foster, Golding, Hsu, Le~Noac'h, Li, McDonell, Muennighoff, Ociepa, Phang, Reynolds, Schoelkopf, Skowron, Sutawika, Tang, Thite, Wang, Wang, and Zou}]{eval-harness}
Leo Gao, Jonathan Tow, Baber Abbasi, Stella Biderman, Sid Black, Anthony DiPofi, Charles Foster, Laurence Golding, Jeffrey Hsu, Alain Le~Noac'h, Haonan Li, Kyle McDonell, Niklas Muennighoff, Chris Ociepa, Jason Phang, Laria Reynolds, Hailey Schoelkopf, Aviya Skowron, Lintang Sutawika, Eric Tang, Anish Thite, Ben Wang, Kevin Wang, and Andy Zou. 2023.
\newblock \href {https://doi.org/10.5281/zenodo.10256836} {A framework for few-shot language model evaluation}.

\bibitem[{Goldstein et~al.(2024)Goldstein, Obeid, Alcaide, Song, and Cheah}]{goldstein2024goldfinch}
Daniel Goldstein, Fares Obeid, Eric Alcaide, Guangyu Song, and Eugene Cheah. 2024.
\newblock Goldfinch: High performance rwkv/transformer hybrid with linear pre-fill and extreme kv-cache compression.
\newblock \emph{arXiv preprint arXiv:2407.12077}.

\bibitem[{Kaddour(2023)}]{kaddour2023minipile}
Jean Kaddour. 2023.
\newblock The minipile challenge for data-efficient language models.
\newblock \emph{arXiv preprint arXiv:2304.08442}.

\bibitem[{Kwon et~al.(2023)Kwon, Li, Zhuang, Sheng, Zheng, Yu, Gonzalez, Zhang, and Stoica}]{kwon2023efficient}
Woosuk Kwon, Zhuohan Li, Siyuan Zhuang, Ying Sheng, Lianmin Zheng, Cody~Hao Yu, Joseph Gonzalez, Hao Zhang, and Ion Stoica. 2023.
\newblock Efficient memory management for large language model serving with pagedattention.
\newblock In \emph{Proceedings of the 29th Symposium on Operating Systems Principles}, pages 611--626.

\bibitem[{Liao and Vargas(2024)}]{liao2024beyond}
Bingli Liao and Danilo~Vasconcellos Vargas. 2024.
\newblock Beyond kv caching: Shared attention for efficient llms.
\newblock \emph{arXiv preprint arXiv:2407.12866}.

\bibitem[{Liu et~al.(2024)Liu, Liu, Pan, He, Haffari, and Zhuang}]{liu2024minicache}
Akide Liu, Jing Liu, Zizheng Pan, Yefei He, Gholamreza Haffari, and Bohan Zhuang. 2024.
\newblock Minicache: Kv cache compression in depth dimension for large language models.
\newblock \emph{arXiv preprint arXiv:2405.14366}.

\bibitem[{Mihaylov et~al.(2018)Mihaylov, Clark, Khot, and Sabharwal}]{mihaylov-etal-2018-suit}
Todor Mihaylov, Peter Clark, Tushar Khot, and Ashish Sabharwal. 2018.
\newblock \href {https://doi.org/10.18653/v1/D18-1260} {Can a suit of armor conduct electricity? a new dataset for open book question answering}.
\newblock In \emph{Proceedings of the 2018 Conference on Empirical Methods in Natural Language Processing}, pages 2381--2391, Brussels, Belgium. Association for Computational Linguistics.

\bibitem[{Mu et~al.(2024)Mu, Wu, Fan, Wang, Li, He, Yang, Xiao, and Zhu}]{mu2024cross}
Yongyu Mu, Yuzhang Wu, Yuchun Fan, Chenglong Wang, Hengyu Li, Qiaozhi He, Murun Yang, Tong Xiao, and Jingbo Zhu. 2024.
\newblock Cross-layer attention sharing for large language models.
\newblock \emph{arXiv preprint arXiv:2408.01890}.

\bibitem[{Rajabzadeh et~al.(2024)Rajabzadeh, Jafari, Sharma, Jami, Kwon, Ghodsi, Chen, and Rezagholizadeh}]{rajabzadeh2024echoatt}
Hossein Rajabzadeh, Aref Jafari, Aman Sharma, Benyamin Jami, Hyock~Ju Kwon, Ali Ghodsi, Boxing Chen, and Mehdi Rezagholizadeh. 2024.
\newblock Echoatt: Attend, copy, then adjust for more efficient large language models.
\newblock \emph{arXiv preprint arXiv:2409.14595}.

\bibitem[{Rajput et~al.(2024)Rajput, Sheng, Owen, and Chiley}]{rajput2024inference}
Shashank Rajput, Ying Sheng, Sean Owen, and Vitaliy Chiley. 2024.
\newblock Inference-friendly models with mixattention.
\newblock \emph{arXiv preprint arXiv:2409.15012}.

\bibitem[{Sakaguchi et~al.(2021)Sakaguchi, Bras, Bhagavatula, and Choi}]{sakaguchi2021winogrande}
Keisuke Sakaguchi, Ronan~Le Bras, Chandra Bhagavatula, and Yejin Choi. 2021.
\newblock Winogrande: An adversarial winograd schema challenge at scale.
\newblock \emph{Communications of the ACM}, 64(9):99--106.

\bibitem[{Shazeer(2019)}]{shazeer2019fast}
Noam Shazeer. 2019.
\newblock Fast transformer decoding: One write-head is all you need.
\newblock \emph{arXiv preprint arXiv:1911.02150}.

\bibitem[{Sheng et~al.(2023)Sheng, Zheng, Yuan, Li, Ryabinin, Chen, Liang, R{\'e}, Stoica, and Zhang}]{sheng2023flexgen}
Ying Sheng, Lianmin Zheng, Binhang Yuan, Zhuohan Li, Max Ryabinin, Beidi Chen, Percy Liang, Christopher R{\'e}, Ion Stoica, and Ce~Zhang. 2023.
\newblock Flexgen: High-throughput generative inference of large language models with a single gpu.
\newblock In \emph{International Conference on Machine Learning}, pages 31094--31116. PMLR.

\bibitem[{Soboleva et~al.(2023)Soboleva, Al-Khateeb, Myers, Steeves, Hestness, and Dey}]{cerebras2023slimpajama}
Daria Soboleva, Faisal Al-Khateeb, Robert Myers, Jacob~R Steeves, Joel Hestness, and Nolan Dey. 2023.
\newblock \href {https://huggingface.co/datasets/cerebras/SlimPajama-627B} {{SlimPajama: A 627B token cleaned and deduplicated version of RedPajama}}.

\bibitem[{Sun et~al.(2024)Sun, Dong, Zhu, Huang, Wang, Ma, Zhang, Wang, and Wei}]{sun2024you}
Yutao Sun, Li~Dong, Yi~Zhu, Shaohan Huang, Wenhui Wang, Shuming Ma, Quanlu Zhang, Jianyong Wang, and Furu Wei. 2024.
\newblock You only cache once: Decoder-decoder architectures for language models.
\newblock \emph{arXiv preprint arXiv:2405.05254}.

\bibitem[{Touvron et~al.(2023)Touvron, Lavril, Izacard, Martinet, Lachaux, Lacroix, Rozi{\`e}re, Goyal, Hambro, Azhar et~al.}]{touvron2023llama}
Hugo Touvron, Thibaut Lavril, Gautier Izacard, Xavier Martinet, Marie-Anne Lachaux, Timoth{\'e}e Lacroix, Baptiste Rozi{\`e}re, Naman Goyal, Eric Hambro, Faisal Azhar, et~al. 2023.
\newblock Llama: Open and efficient foundation language models.
\newblock \emph{arXiv preprint arXiv:2302.13971}.

\bibitem[{Welbl et~al.(2017)Welbl, Liu, and Gardner}]{welbl-etal-2017-crowdsourcing}
Johannes Welbl, Nelson~F. Liu, and Matt Gardner. 2017.
\newblock \href {https://doi.org/10.18653/v1/W17-4413} {Crowdsourcing multiple choice science questions}.
\newblock In \emph{Proceedings of the 3rd Workshop on Noisy User-generated Text}, pages 94--106, Copenhagen, Denmark. Association for Computational Linguistics.

\bibitem[{Wolf et~al.(2020)Wolf, Debut, Sanh, Chaumond, Delangue, Moi, Cistac, Rault, Louf, Funtowicz, Davison, Shleifer, von Platen, Ma, Jernite, Plu, Xu, Le~Scao, Gugger, Drame, Lhoest, and Rush}]{wolf-etal-2020-transformers}
Thomas Wolf, Lysandre Debut, Victor Sanh, Julien Chaumond, Clement Delangue, Anthony Moi, Pierric Cistac, Tim Rault, Remi Louf, Morgan Funtowicz, Joe Davison, Sam Shleifer, Patrick von Platen, Clara Ma, Yacine Jernite, Julien Plu, Canwen Xu, Teven Le~Scao, Sylvain Gugger, Mariama Drame, Quentin Lhoest, and Alexander Rush. 2020.
\newblock \href {https://doi.org/10.18653/v1/2020.emnlp-demos.6} {Transformers: State-of-the-art natural language processing}.
\newblock In \emph{Proceedings of the 2020 Conference on Empirical Methods in Natural Language Processing: System Demonstrations}, pages 38--45, Online. Association for Computational Linguistics.

\bibitem[{Wu and Tu(2024)}]{wu-tu-2024-layer}
Haoyi Wu and Kewei Tu. 2024.
\newblock \href {https://doi.org/10.18653/v1/2024.acl-long.602} {Layer-condensed {KV} cache for efficient inference of large language models}.
\newblock In \emph{Proceedings of the 62nd Annual Meeting of the Association for Computational Linguistics (Volume 1: Long Papers)}, pages 11175--11188, Bangkok, Thailand. Association for Computational Linguistics.

\bibitem[{Xiao et~al.(2024)Xiao, Tian, Chen, Han, and Lewis}]{xiao2024efficient}
Guangxuan Xiao, Yuandong Tian, Beidi Chen, Song Han, and Mike Lewis. 2024.
\newblock \href {https://openreview.net/forum?id=NG7sS51zVF} {Efficient streaming language models with attention sinks}.
\newblock In \emph{The Twelfth International Conference on Learning Representations}.

\bibitem[{Zellers et~al.(2019)Zellers, Holtzman, Bisk, Farhadi, and Choi}]{zellers-etal-2019-hellaswag}
Rowan Zellers, Ari Holtzman, Yonatan Bisk, Ali Farhadi, and Yejin Choi. 2019.
\newblock \href {https://doi.org/10.18653/v1/P19-1472} {{H}ella{S}wag: Can a machine really finish your sentence?}
\newblock In \emph{Proceedings of the 57th Annual Meeting of the Association for Computational Linguistics}, pages 4791--4800, Florence, Italy. Association for Computational Linguistics.

\bibitem[{Zhang et~al.(2024)Zhang, Zeng, Wang, and Lu}]{zhang2024tinyllama}
Peiyuan Zhang, Guangtao Zeng, Tianduo Wang, and Wei Lu. 2024.
\newblock Tinyllama: An open-source small language model.
\newblock \emph{arXiv preprint arXiv:2401.02385}.

\bibitem[{Zhang et~al.(2023)Zhang, Sheng, Zhou, Chen, Zheng, Cai, Song, Tian, Re, Barrett, Wang, and Chen}]{zhang2023ho}
Zhenyu Zhang, Ying Sheng, Tianyi Zhou, Tianlong Chen, Lianmin Zheng, Ruisi Cai, Zhao Song, Yuandong Tian, Christopher Re, Clark Barrett, Zhangyang Wang, and Beidi Chen. 2023.
\newblock \href {https://openreview.net/forum?id=RkRrPp7GKO} {H2o: Heavy-hitter oracle for efficient generative inference of large language models}.
\newblock In \emph{Thirty-seventh Conference on Neural Information Processing Systems}.

\bibitem[{Zuhri et~al.(2024)Zuhri, Adilazuarda, Purwarianti, and Aji}]{zuhri2024mlkv}
Zayd Muhammad~Kawakibi Zuhri, Muhammad~Farid Adilazuarda, Ayu Purwarianti, and Alham~Fikri Aji. 2024.
\newblock Mlkv: Multi-layer key-value heads for memory efficient transformer decoding.
\newblock \emph{arXiv preprint arXiv:2406.09297}.

\end{thebibliography}

\appendix

\section{Model and Training Details}
\label{apx:details}

Table \ref{tab:configurations} and \ref{tab:hyperparameters} show the model configurations and training details for Section \ref{sec:experiments}.
The configuration of the 1.1B model follows that of TinyLlama \cite{zhang2024tinyllama}.
We use the MiniPile \cite{kaddour2023minipile} (licensed under MIT) and SlimPajama \cite{cerebras2023slimpajama} (various licenses depending on the data source) as our datasets. Our use of the datasets is consistent with their intended use.

\begin{table}[tbh]
  \centering
  \begin{tabular}{lccc}
    \toprule
    Model Size & 110M & 1.1B \\
    \midrule
    Hidden Size & 768 & 2048 \\
    Intermediate Size & 2048 & 5632 \\
    Max Trained Length & 1024 & 2048 \\
    \# Layers & 12 & 22 \\
    \# Attention Heads & 12 & 32 \\
    \# KV Heads & 6 & 4 \\
    \bottomrule
  \end{tabular}
  \caption{Model configurations.}
  \label{tab:configurations}
\end{table}

\begin{table*}[tbh]
  \centering
  \begin{tabular}{lccc}
    \toprule
    Section & \multicolumn{2}{c}{\ref{sec:small-experiments}} & \ref{sec:large-experiments} \\
    \cmidrule(l){2-4} 
    Model Size & 110M & 1.1B & 1.1B \\
    \midrule
    Max LR & 6.75e-4 & 3e-4 & 4e-4 \\
    Min LR & 0 & 0 & 4e-5 \\
    LR Scheduler & \multicolumn{3}{c}{cosine} \\
    Optimizer & \multicolumn{3}{c}{AdamW} \\
    $\beta1$ & \multicolumn{3}{c}{0.9} \\
    $\beta2$ & 0.999 & 0.999 & 0.95 \\
    Warmup Ratio & 0.015 & 0.015 & 200 steps \\
    Weight Decay & \multicolumn{3}{c}{0.1} \\
    Gradient Clipping & \multicolumn{3}{c}{1.0} \\
    Batch Size (tokens) & 32K & 256K & 2M \\
    Epochs & 2 & 1 & 100B tokens \\
    GPU & RTX 3090x1 & A100x8 & A800x128 \\
    \bottomrule
  \end{tabular}
  \caption{Training details.}
  \label{tab:hyperparameters}
\end{table*}

\begin{table*}[tbh]
  \centering
  \small
  \begin{tabular}{clcccccccc}
    \toprule
    \# KV Layers & Model & Hellaswag & Obqa & WG & ARC-c & ARC-e & BoolQ & PIQA & SciQ \\
    \midrule
    22 & Standard Transformer & 44.58 & 30.2 & 50.99 & 25.00 & 46.38 & 60.46 & 68.93 & 74.8 \\
    \midrule
    \multirow{4}{*}{11} & Pizza-Bottom & 44.20 & 29.4 & 51.93 & 25.00 & 46.55 & 59.51 & 68.28 & 72.1 \\
    & Lasagna-Bottom & 43.43 & 30.8 & 50.51 & 24.49 & 44.61 & 59.24 & 69.21 & 71.5 \\
    & Sandwich-Top & 44.74 & 31.0 & 51.70 & 24.83 & 46.38 & 61.38 & 67.90 & 72.5 \\
    & Sandwich-Middle & 44.22 & 31.0 & 52.01 & 24.49 & 44.86 & 58.62 & 68.39 & 70.7 \\
    \midrule
    \multirow{4}{*}{7} & Pizza-Bottom & 42.79 & 30.0 & 52.25 & 24.74 & 45.37 & 56.82 & 68.61 & 71.0 \\
    & Lasagna-Bottom & 42.86 & 31.6 & 53.43 & 25.17 & 45.79 & 59.79 & 68.22 & 69.1 \\
    & Sandwich-Top & 43.88 & 30.0 & 52.83 & 25.68 & 43.73 & 61.07 & 67.57 & 69.5 \\
    & Sandwich-Middle & 43.84 & 30.0 & 51.77 & 25.68 & 45.50 & 60.73 & 68.77 & 68.1 \\
    \midrule
    \multirow{4}{*}{3} & Pizza-Bottom & 40.21 & 30.4 & 51.93 & 24.06 & 43.18 & 58.65 & 67.13 & 68.4 \\
    & Lasagna-Bottom & 41.76 & 28.0 & 52.25 & 26.02 & 44.36 & 57.28 & 67.90 & 69.8 \\
    & Sandwich-Top & 42.14 & 30.2 & 49.80 & 24.91 & 43.39 & 61.47 & 66.97 & 68.9 \\
    & Sandwich-Middle & 43.43 & 31.0 & 51.70 & 24.40 & 44.95 & 59.57 & 68.17 & 67.3 \\
    \bottomrule
  \end{tabular}
  \caption{Detailed downstream task results of 1.1B models trained on the Slimpajama dataset.}
  \label{tab:downstream}
\end{table*}

\section{Throughput at Different Batch Sizes}
\label{apx:batchsize}

\begin{figure}[tbh]
    \centering
    \includegraphics[trim = 0mm 0mm 0mm 0mm, clip, width=\columnwidth]{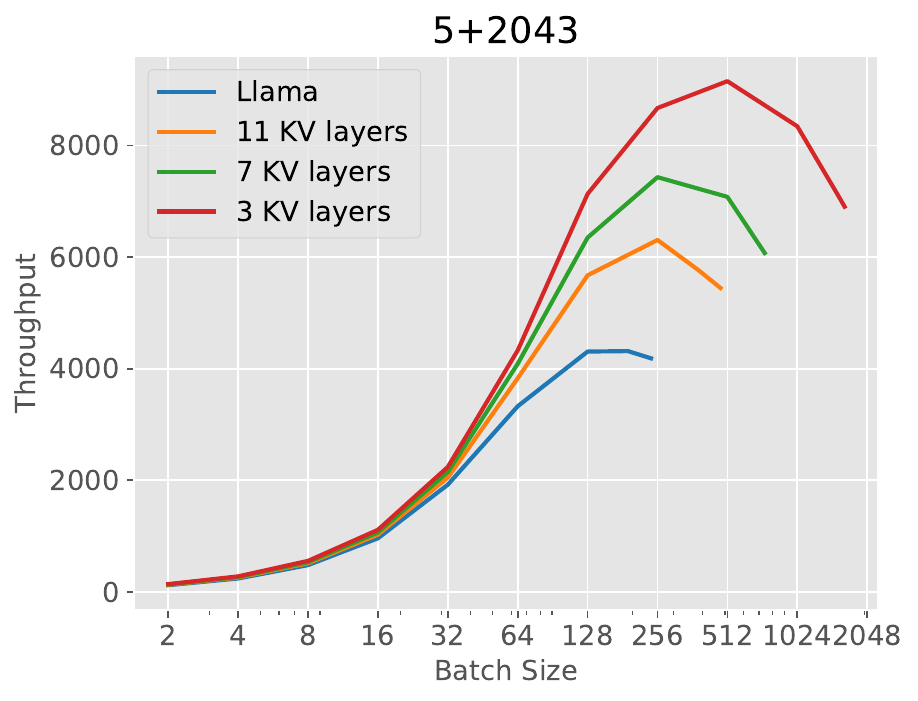}
    \caption{Throughput of 1.1B models at different batch sizes on an RTX 3090 (24GB) GPU with a prompt length of 5 and a generation length of 2043.}
    \label{fig:batchsize}
\end{figure}

Figure \ref{fig:batchsize} reports the generation throughput of the standard Llama and the nine configurations with different numbers of KV layers at different batch sizes. The highest point of each curve indicates the maximum throughput of the model, which has been shown in Figure \ref{fig:experiment}(a), and the rightmost point indicates the maximum batch size. It can be seen that, at any given batch size, the throughput of the nine configurations is higher than the baseline throughput and increases as the number of KV layers decreases.

\section{Detailed Downstream Task Results}
\label{apx:downstream}

Table \ref{tab:downstream} reports the accuracy of each downstream task of the models in Section \ref{sec:large-experiments}.

\section{Initializing with Pre-trained Models}
\label{apx:initialize}

Instead of training from scratch, we can initialize the standard Llama and the nine configurations with pre-trained models to get better performance.
We follow the uptraining scheme of MLKV \cite{zuhri2024mlkv}.
For each KV layer, we initialize the weights $W_K,W_V$ with the averaged weights of all layers whose queries are paired with its KVs.
We use the TinyLlama checkpoint trained on 2.5T tokens to initialize the models with 1.1B parameters.
The training details are the same as in Section \ref{sec:small-experiments}.

\begin{figure}[tbh]
    \centering
    \begin{subfigure}[b]{\columnwidth}
        \centering
        \includegraphics[trim = 0mm 150mm 0mm 0mm, clip, width=\columnwidth]{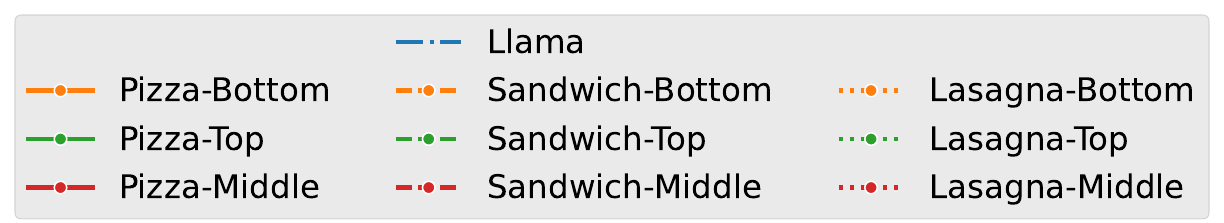}
    \end{subfigure}
    \begin{subfigure}[b]{\columnwidth}
        \centering
        \includegraphics[trim = 0mm 0mm 0mm 47mm, clip, width=\columnwidth]{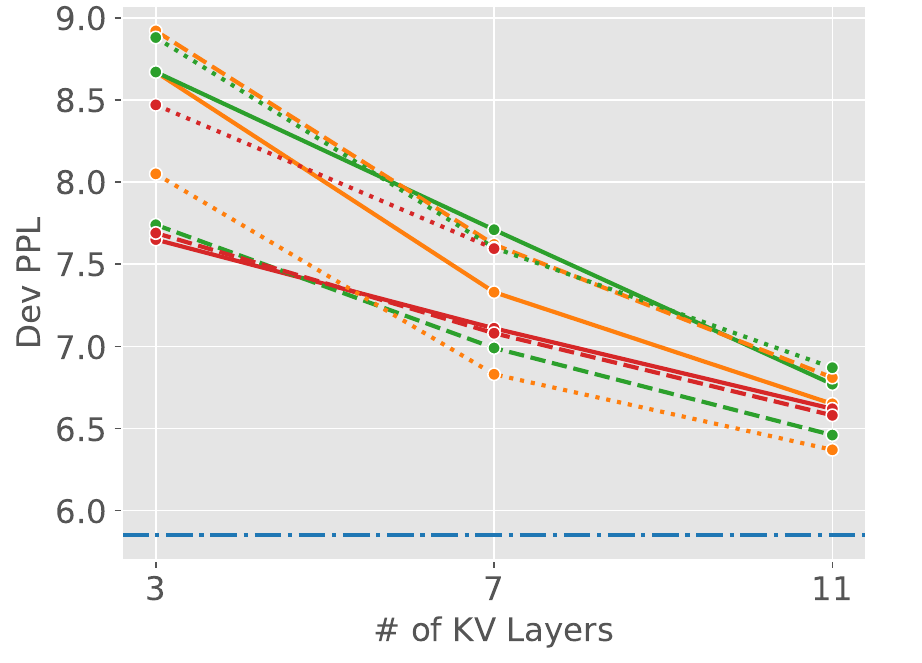}
    \end{subfigure}
    \caption{Perplexity on the Minipile dataset of 1.1B models initialized with converted Tinyllama-2.5T weights.}
    \label{fig:initialize}
\end{figure}

Figure \ref{fig:initialize} reports the perplexity.
It can be seen that all models achieve better performance, compared to training from scratch.
The lasagna-bottom configuration performs best when retaining 11 and 7 KV layers, but was surpassed by some top and middle configurations when retaining 3 KV layers.
Notice that for the top and middle positioning, we drop the attention of each token to itself and therefore differ from the standard transformer.
In future work, we will try to make up for this gap by specially computing the attention of each token to itself, and we hope to get a better performance.

\section{More Options for Target Layer Positioning}
\label{apx:quarter}

In addition to positioning the target layer at the top, bottom, and middle, we also consider the quarter and three-quarter, and name the corresponding configurations as middle-1/4 and middle-3/4.
We train the new configurations with 1.1B parameters.
The training details are the same as in Section \ref{sec:small-experiments}.

\begin{figure}[tbh]
    \centering
    \begin{subfigure}[b]{\columnwidth}
        \centering
        \includegraphics[trim = 0mm 142mm 0mm 0mm, clip, width=0.92\columnwidth]{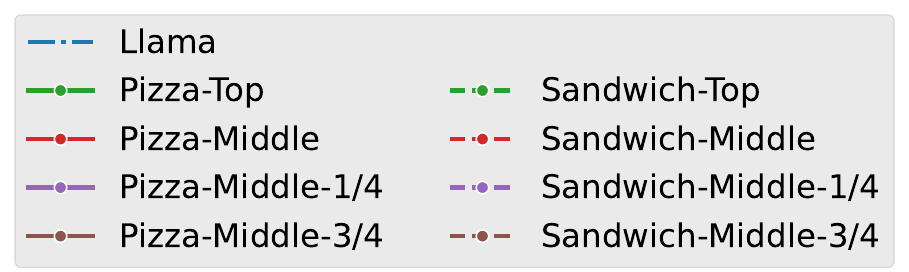}
    \end{subfigure}
    \begin{subfigure}[b]{\columnwidth}
        \centering
        \includegraphics[trim = 0mm 0mm 0mm 47mm, clip, width=\columnwidth]{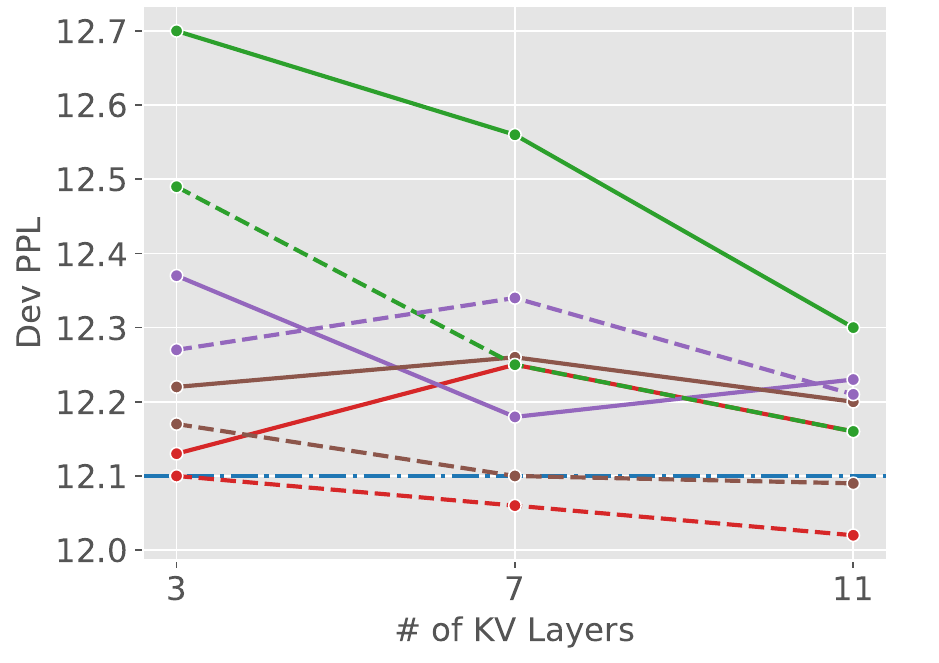}
    \end{subfigure}
  \caption{Perplexity on the Minipile dataset of 1.1B models with more options for target layer positioning.}
  \label{fig:quarter}
\end{figure}

Figure \ref{fig:quarter} reports the perplexity.
We omit lasagna configurations because there are not enough layers in each group to distinguish between different target layer positions.
It can be seen that the performance of the middle-1/4 and middle-3/4 configurations mainly lies between the top and middle configurations.

\end{document}